\documentclass[
reprint,
superscriptaddress,
 amsmath,amssymb,
 aps,
pre,
]{revtex4-2}

\usepackage{graphicx}
\usepackage{dcolumn}
\usepackage{bm}
\usepackage{xcolor}
\usepackage{multirow}
\usepackage{hyperref}
\hypersetup{colorlinks=true,linkcolor=blue,citecolor=blue}
\usepackage{array}

\begin{document}

\title{Learning unidirectional coupling using echo-state network}

\author{Swarnendu Mandal}
\affiliation{Central University of Rajasthan, Ajmer, Rajasthan, India - 305817}

\author{Manish Dev Shrimali}
\email{shrimali@curaj.ac.in}
\affiliation{Central University of Rajasthan, Ajmer, Rajasthan, India - 305817}

\begin{abstract}
Reservoir Computing has found many potential applications in the field of complex dynamics. In this article, we exploit the exceptional capability of the echo-state network (ESN) model to make it learn a unidirectional coupling scheme from only a few time series data of the system. We show that, once trained with a few example dynamics of a drive-response system, the machine is able to predict the response system's dynamics for any driver signal with the same coupling. Only a few time series data of an $A-B$ type drive-response system in training is sufficient for the ESN to learn the coupling scheme. After training even if we replace drive system $A$ with a different system $C$, the ESN can reproduce the dynamics of response system $B$ using the dynamics of new drive system $C$ only.
\end{abstract}

\maketitle

\section{Introduction}\label{sec:intro}

Application of machine learning in studying dynamical systems has emerged as a very successful area of research in recent time \cite{choudhary2020physics,han2021adaptable,meiyazhagan2021model,choudhary2021forecasting,miller2020scaling}. Not only machine learning has helped to understand complex dynamics better, but also the analysis of the complex dynamics involved in many machine learning algorithms has enriched the field \cite{verzelli2021learn,ceni2020echo,aljadeff2016low,gallicchio2017echo}. Such a computational framework bridging these two research areas, called reservoir computing (RC), is recently gaining overwhelming attention of researchers \cite{tanaka2019recent,nakajima2020physical}. Reservoir computing is a recurrent neural network (RNN) based deep learning architecture derived from two independently proposed concepts, Echo State Network (ESN) by Jaeger \cite{jaeger2001echo} and Liquid State Machine (LSM) by Maass \cite{maass2002real}. The reason for its overarching success in studying dynamical systems lies behind its computational cost efficiency and exceptional accuracy in temporal data processing \cite{lukovsevivcius2009reservoir,lukovsevivcius2012reservoir}. Thus, reservoir computing has found many potential applications in studying complex dynamics in terms of time series prediction \cite{pathak2018model}, attractor reconstruction\cite{lu2018attractor}, identifying the presence of multistability in a system \cite{rohm2021model}, basin prediction \cite{roy2022model}, prediction of dynamical transitions \cite{xiao2021predicting} and many more \cite{ghosh2021reservoir,roy2022role,krishnagopal2020separation}. Even a few recent developments have shown it to be able to approximate the whole system's behavior learning from a few local example dynamics only  \cite{kim2021teaching}. Another dimension of research considers a dynamical system as the main computing substrate, the reservoir, to perform machine learning tasks using reservoir computing technique  \cite{appeltant2011information,coulombe2017computing,choi2019critical,mandal2021achieving,jensen2017reservoir,mandal2022machine}. Thus, RC becomes more relevant in the context of studying complex dynamics as it provides a platform to utilize natural information processing capacity of a dynamical system.

In this article, we use the echo-state network (ESN) approach of RC to explore another potential application in dynamical system research. This conventional RC architecture consists of a random recurrent neural network usually driven by a one or multi-dimensional temporal input signal \cite{jaeger2001echo,lukovsevivcius2009reservoir}. The temporal states generated by the activation of neurons are used for linear regression to find the output. The main advantage of ESN over other conventional RNN architectures is that it does not require training the internal weights. So the main computing layer, the reservoir remains unchanged during training. From the viewpoint of dynamics, the reservoir layer acts as a dynamical system that internalizes the input signal into its dynamics and transforms it into a very high-dimensional spatio-temporal pattern. Finally, the output connections are trained to convert this pattern to the required output using the training dataset.

\begin{figure}
    \centering
    \includegraphics[scale=0.3]{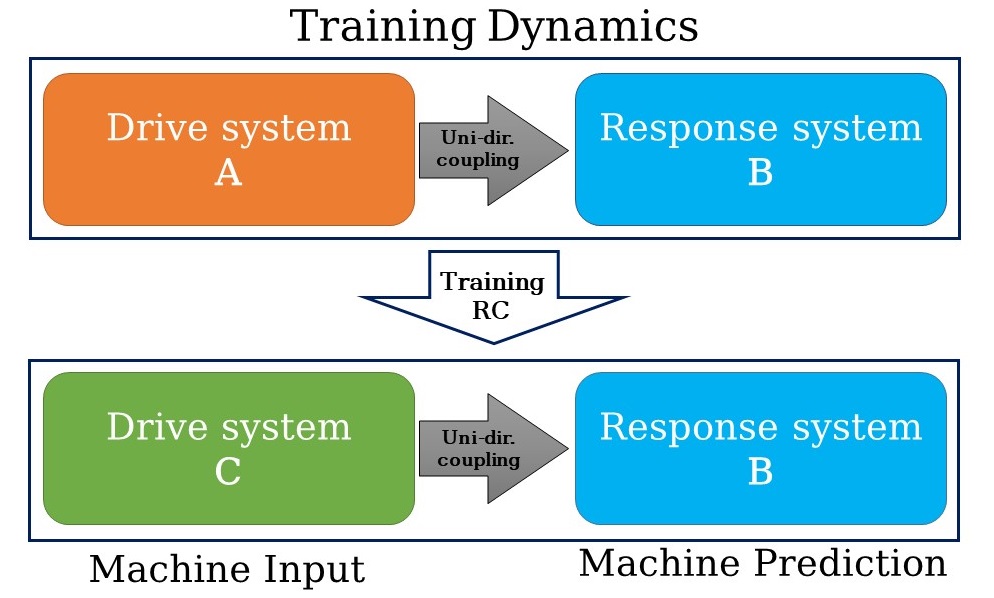}
    \caption{Schematic for learning of unidirectional coupling by a reservoir computer.}
    \label{fig:sch_work}
\end{figure}

Now, in the context of research in the field of complex dynamics, the study of various emergent behavior in coupled oscillators have always been of great importance. Presence of coupled behavior in variety of natural systems like physical \cite{aronson1990amplitude}, chemical \cite{bar1985stability}, biological \cite{strogatz1993coupled}, neuronal \cite{steur2009semi} and many more, has sought attention of many researchers. Several models considering various coupling scheme has been shown to produce different dynamical behaviors like oscillation quenching \cite{sharma2012amplitude}, synchronization \cite{li2012synchronization}, chimera states \cite{abrams2004chimera} even induced multistability \cite{kim1997multistability}. One of such interacting oscillator systems exhibits a one-way coupling scheme leading to a unique type of system, often called as drive-response system or master-slave system \cite{pecora1991driving}. Starting from exploring the effect of a chaotic drive signal on a stable autonomous system \cite{pecora1991driving}, this kind of causally connected systems has been extensively studied over time in the field of complex dynamics \cite{uchida2004consistency,sharma2014controlling,agrawal2013driving,ujjwal2016driving}. Drive-response system has played a crucial role in understanding and exploration of many important concepts of complex dynamical behavior like theory of synchronization \cite{pecora1990synchronization}. The synchronization of these unidirectinally coupled systems has helped establishing many important ideas regarding coupled systems \cite{tresser1995master,mei2013finite} and also has been applied in the field of cryptography \cite{li2009projective}. Moreover, researchers have even used these systems for analysing the emerging chimera states in a complex network \cite{botha2018analysis}. However, despite having this much significance very few studies have focused on the key feature of such systems which is the causal relation between the coupled pair. In this context, an earlier study has exploited some interdependence measures to identify the directionality (which system is driving and which one is following) of a drive-response system \cite{quiroga2000learning}.

In this article, we use a reservoir computer to learn the unidirectional coupling scheme from a set of dynamics of such systems as shown by the schematic in Fig.~\ref{fig:sch_work}. We show that, after learning from a few example dynamics of a drive-response system, the ESN can predict the response system's behavior for a completely new drive signal, even from a previously unseen system, given the coupling is the same. This implies that the machine is able to learn the coupling scheme, the way drive system affects response system's dynamics. We consider a system with unidirectional coupling to train the ESN with its dynamics and show the result of the machine's prediction of response dynamics for a completely different drive signal with the same coupling.

Many natural systems may have external independent driving factors influencing the original dynamics. Most of the time it is difficult to mathematically model such spontaneous systems. Our proposed idea can predict the modified dynamics from the information of external drive signals by learning only from data. Considering the abundance in the natural occurrence of such systems, this reservoir computing scheme may be of potential applications. However, this scheme works only for a scalar drive signal. Use of multi-dimensional drive signal leads the reservoir to work as an observer (observer effect \cite{lu2017reservoir}). This results in an ambiguity during the prediction with multi-dimensional drive signal from a different system.

Rest of the article is organized in the following way. Sec.~\ref{sec:esn} discusses the general architecture of ESN used for this work. Next, in Sec.~\ref{sec:res}, we discuss the results considering two different systems with unidirectional coupling. For each case we consider a drive-response system to train the machine and then predict the response system's behaviour with a different drive system. Finally in Sec.~\ref{sec:disc}, we conclude our findings and discuss about importance and scope of this work.

\section{Echo-State Network}\label{sec:esn}
\begin{figure}
    \centering
    \includegraphics[scale=0.25]{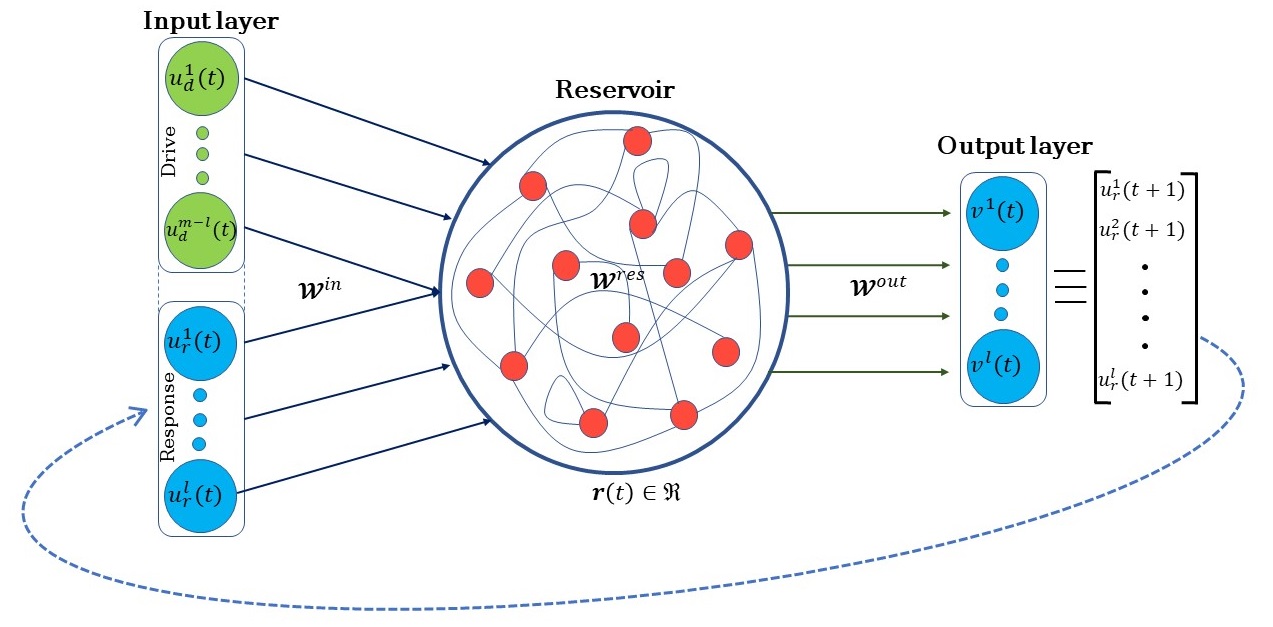}
    \caption{Schematic of ESN architecture.}
    \label{fig:sch_ESN}
\end{figure}

In this article, we implement the ESN architecture originally proposed by Jaeger \cite{jaeger2001echo}. Conventionally, this kinds of reservoir computer consist of three components as depicted in Fig.~\ref{fig:sch_ESN}. First, the {\em input layer} which contains an $(m\times N)$-dimensional weighted matrix $\mathcal{W}^{in}$, called input connection matrix. It takes an $m$-dimensional signal $\mathbf{u}(t)$ as input and converts into an $N$-dimensional signal. The elements of $\mathcal{W}^{in}$ are randomly chosen from range $[-\sigma,\sigma]$, where $\sigma$ is a hyper-parameter for the scheme. Second, the {\em reservoir layer} which consists of a network of $N$ nodes, characterised by an $(N\times N)$-dimensional weighted matrix $\mathcal{W}^{res}$. This is again a random sparse matrix with spectral radius $\rho$ which acts as a hyper-parameter to be optimized. All the nodes in this layer have individual states collectively represented as the vector $\bm{r}(t)$. The dynamics of this reservoir driven by the input signal are defined by state updating map, as
\begin{equation}\label{eq:st_upd}
    \bm{r}(t+1) = (1-\alpha)\bm{r}(t) + \alpha~\mathrm{tanh}[\mathcal{W}^{res}\cdot \bm{r}(t) + \mathcal{W}^{in} \cdot \bm{u}(t)],
\end{equation}
where, $\alpha$ is a hyper-parameter called leaking parameter which takes a value between $0$ to $1$. The third component, {\em output layer} evaluates the $l$-dimensional output signal $\bm{v}(t)$ from reservoir state $\bm{r}(t)$. This layer is made up of an $(N\times l)$-dimensional output connection matrix $\mathcal{W}^{out}$, which is the only trainable part of ESN architecture.

Conventionally, the dimensions of input signal $(m)$ and output signal $(l)$ are taken sufficiently lower than the number of reservoir nodes i.e. $m\sim l<<N$. For our case of predicting response systems behavior, we consider all the state variables of a drive-response system in the input and immediate next state of the response system only as the output. Thus $m$ is the total number of state variables of the coupled system and $l$ is the number of state variables in the response system.

In the training phase, after generating $\mathcal{W}^{in}$ and $\mathcal{W}^{res}$ we run the reservoir following Eq.~(\ref{eq:st_upd}) using the $m$-dimensional input time series data $\bm{u}(t)$. The input connection matrix is generated in such a way that information from each of the input channels is equally distributed over all reservoir nodes. Thus, any one input dimension is connected to $N/m$ number of nodes in the reservoir network. After removing enough transient evolution, the dynamics of the reservoir states $\bm{r}(t)$ is stored as $\bm{\tilde{r}}(t)$ in a $(N\times T)$-dimensional matrix $\Re$, where $T$ is the number of time steps of input signal used for training after transient steps. The store reservoir states $\bm{\tilde{r}}(t)$ are generated such a way that the odd row element is taken from odd row elements of $\bm{r}(t)$ and the even row elements of $\bm{\tilde{r}}(t)$ are the even row elements of $\bm{r}^2(t)$ \cite{pathak2018model}. So if $\bm{r}(t) = [r_1(t)~r_2(t)~r_3(t).....~r_N(t)]^T$ and $\bm{\tilde{r}}(t) = [\tilde{r}_1(t)~\tilde{r}_2(t)~\tilde{r}_3(t).....~\tilde{r}_N(t)]^T$ then

\begin{equation}
\label{eq:res_st}
\tilde{r}_i(t)=
    \begin{cases}
        r_i(t) & \text{for odd } i \\
        r^2_i(t) & \text{for even } i.
    \end{cases} 
\end{equation}

Now, we use the response system's state variable at very next time step as the label (or target output) to train the ESN. So if the input signal $\bm{u}(t)$ consists of the state variables drive system $\bm{u}_d(t)$ and state variables response system $\bm{u}_r(t)$ as $\bm{u}(t) = [\bm{u}_d(t)\quad \bm{u}_r(t)]^T$, then the labels used for training is $\bm{u}_r(t+1)$ corresponding to input $\bm{u}(t)$. So in the training phase, after generating the reservoir state matrix $\Re$ one can readily evaluate the output connection matrix $\mathcal{W}^{out}$ using the linear transformation equation between output and reservoir states

\begin{equation}
\label{eq:res_out}
    \mathcal{U} = \mathcal{W}^{out}\cdot \Re,
\end{equation}
where $\mathcal{U}$ is the $l\times T$ dimensional matrix containing target output corresponding to training input. Now, the set of optimal $\mathcal{W}^{out}$ is evaluated using Ridge regression minimising the difference between $\mathcal{U}$ and $\mathcal{W}^{out}\cdot \Re$ such that $(\mathcal{U} - \mathcal{W}^{out}\cdot \Re) \approx 0$. This leads to 

\begin{equation}
    \mathcal{W}^{out}=\mathcal{U}\Re^T(\Re\Re^T + \beta I)^{-1},
\end{equation}
where, $\beta$ is called the regularization parameter which controls the over-fitting. This also can be treated as a hyper-parameter subject to optimization. Once we have $\mathcal{W}^{out}$ in hand, evaluated using training data set, machine can predict the required output from any given input using Eq.~(\ref{eq:res_out}). Particularly for our case, ESN is now ready to predict the response system's dynamics using the corresponding drive signal.

In the prediction phase, we generate the drive system's data by solving the model, and the response system's dynamics are generated by ESN working in a close-loop configuration. We run the reservoir using the same update rule define in Eq.~(\ref{eq:st_upd}) using an input $\bm{\tilde{u}}(t)$ as

\begin{equation}
    \label{eq:st_upd_pred}
    \bm{r}(t+1) = (1-\alpha)\bm{r}(t) + \alpha~\mathrm{tanh}[\mathcal{W}^{res}\cdot \bm{r}(t) + \mathcal{W}^{in} \cdot \bm{\tilde{u}}(t)],
\end{equation}
where, $\bm{\tilde{u}}(t)$ consists of the drive signal $\bm{u}_d(t)$ evaluated from the model and the output of ESN $\bm{v}(t)$ as the response dynamics i.e. $\bm{\tilde{u}}(t) = [\bm{u}_d(t)\quad \bm{v}(t)]$ (see Fig.~\ref{fig:sch_ESN}). Here, $\bm{v}(t)$ is calculated using Eq.~(\ref{eq:res_out}) as

\begin{equation}
    \bm{v}(t) = \mathcal{W}^{out}\cdot \bm{\tilde{r}}(t),
\end{equation}

where, $\bm{\tilde{r}}(t)$ is evaluated using reservoir states $\bm{r}(t)$ in the similar fashion described in Eq.~(\ref{eq:res_st}). Thus starting from an initial state of the response system, ESN can produce the response system's dynamics using the drive signal working in an iterative process. Here, at the beginning of the prediction phase, we ignore a few reservoir iteration states to avoid the effect of reservoir initialization. This is termed as `{\em warm up}' period for the reservoir.  For all these iteration steps, we use the same initial value of input $\bm{\tilde{u}}(t_0)$, and then we start the ESN to work in a closed-loop configuration.

The overall procedure involves many hyper-parameters, like the number of reservoir nodes ($N$), range of weights in input connection matrix $\mathcal{W}^{in}$ ($\sigma$), spectral radius of the reservoir ($\rho$), the leaking parameter ($\alpha$) and regularization parameter ($\beta$) in Ridge regression. We have observed that the improvement of accuracy of the scheme saturates at a point with the increment of reservoir nodes number ($N$). So, considering the fact that a greater number of nodes requires increased computational power, we have chosen $N=1200$ for all the numerical simulations for this work. A greater value of $N$ yields an insignificant accuracy improvement. We use the Bayesian optimization technique to optimize the rest of the hyper-parameters according to the tasks performed \cite{yperman2016bayesian,griffith2019forecasting,xiao2021predicting}. The loss function for this is defined as the root-mean-square error of a finite predicted time series from its actual target, averaged over multiple reservoir realizations. The optimization is implemented with Gaussian processes (GP) surroget model using \texttt{gp\_minimize} function of \texttt{skopt} python package.

\section{Results}\label{sec:res}

To apply the ESN architecture described in Sec.~\ref{sec:esn}, we consider two different drive-response systems namely, system {\em A-B} and system {\em C-B}, say. We present our results in two ways. First, we show the prediction performance with the same drive system as training but at different parameters. Then, the result of response prediction with a completely different drive system. For example, if we train the machine with system {\em A-B}, then we present the result of response ({\em B}) prediction for the same drive system ({\em A}) as well as for a new drive system ({\em C}). To show the generality of our scheme, we present the results of the machine's performance while trained with system {\em C-B} too. Then system {\em A} works as a new drive system for prediction of {\em B}.

Here, we consider two unidirectionally coupled systems to apply our scheme. One is a non-identical R{\"o}ssler-R{\"o}ssler ({\em A-B}) system, given as

\begin{equation}\label{eq:ros-ros}
    \begin{aligned}
    \dot{x}_d &= -y_d - z_d\\
    \dot{y}_d &= x_d + a'_{ros} y_d\\
    \dot{z}_d &= b'_{ros} + z_d(x_d-c'_{ros})\\
    \dot{x}_r &= -y_r - z_r\\
    \dot{y}_r &= x_r + a_{ros} y_r\\
    \dot{z}_r &= b_{ros} + z_r(x_r-c_{ros}) + \epsilon(z_d - z_r).
    \end{aligned}
\end{equation}
For all simulations we have taken $a'_{ros} = a_{ros} = 0.2$, $b'_{ros} = b_{ros} = 0.2$ and $c_{ros} = 5.7$.
 Different R{\"o}ssler drive signals used for training and prediction are generated by varying values of $c'_{ros} (\neq c_{ros})$. But all the parameters are chosen to configure both the drive and response system in the chaotic regime.

Another drive-response, we consider to be a Lorenz-R{\"o}ssler ({\em C-B}) system given as

\begin{equation}\label{eq:lor-ros}
    \begin{aligned}
    \dot{x}_d &= \sigma_{lor}(y_r-x_r)\\
    \dot{y}_d &= \rho_{lor} x_r - y_r - x_r z_r\\
    \dot{z}_d &= x_r y_r - \beta_{lor} z_r\\
    \dot{x}_r &= -y_r - z_r\\
    \dot{y}_r &= x_r + a_{ros} y_r\\
    \dot{z}_r &= b_{ros} + z_r(x_r-c_{ros}) + \epsilon(z_d - z_r).
    \end{aligned}
\end{equation}
Here, the response system's parameter are taken same as in Eq.~(\ref{eq:ros-ros}) and for the drive signal,  $\sigma_{lor} = 10$ and $\beta_{lor} = 8/3$ are kept fixed.
Different Lorenz drive signal used for training and prediction are generated by varying values of $\rho_{lor}$.

\begin{figure*}
    \centering
    \includegraphics{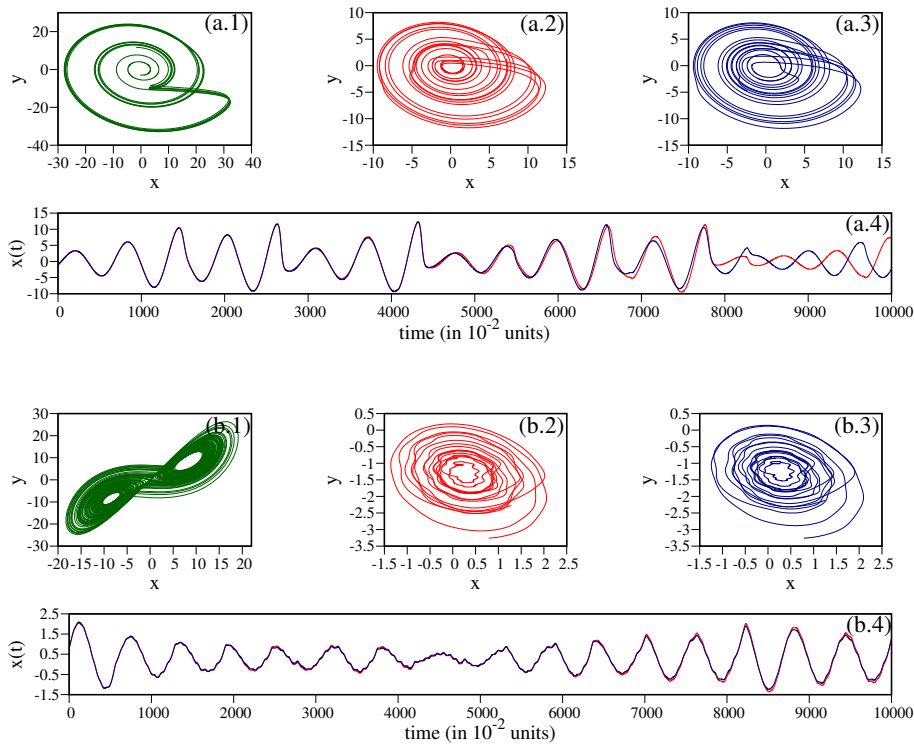}
    \caption{Results of ESN's prediction while trained with R{\"o}ssler-R{\"o}ssler system. (a.1) R{\"o}ssler drive signal for $c'_{ros} = 18$, (a.2 - a.3) respectively, the target and machine prediction for response dynamics corresponding to (a.1), (a.4) comparison of target (red line) and predicted (blue line) time series with $\rm{log_{10}(NRMSE)} = -3.96$.
    (b.1) Lorenz drive signal for $\rho_{lor} = 38$, (b.2 - b.3) respectively, the target and machine prediction for response dynamics corresponding to (b.1), (b.4) comparison of target (red line) and predicted (blue line) time series with $\rm{log_{10}(NRMSE)} = -4.79$.}
    \label{fig:res1}
\end{figure*}

In the first case, we start with training the machine with the R{\"o}ssler-R{\"o}ssler system described in Eq.~(\ref{eq:ros-ros}). The ESN is trained with only three time series for three different values of drive signal parameter $c'_{ros}$ such that $c'_{ros}\in [5.0,10.0,15.0]$. This is enough for the machine to predict the response system's dynamics for any given drive signal keeping the coupling the same. Fig.~\ref{fig:res1} shows the results of its prediction for a R{\"o}ssler drive signal for $c'_{ros} = 18.0$ as well as with a Lorenz drive signal as in Eq.~(\ref{eq:lor-ros}) for $\rho_{lor} = 38$. For this simulation, the Bayesian optimal hyper-parameters are found to be $\sigma = 0.0639$, $\rho = 0.5057$, $\alpha = 0.6057$ and $\beta = 4.7487\times 10^{-5}$.

\begin{figure*}
    \centering
    \includegraphics{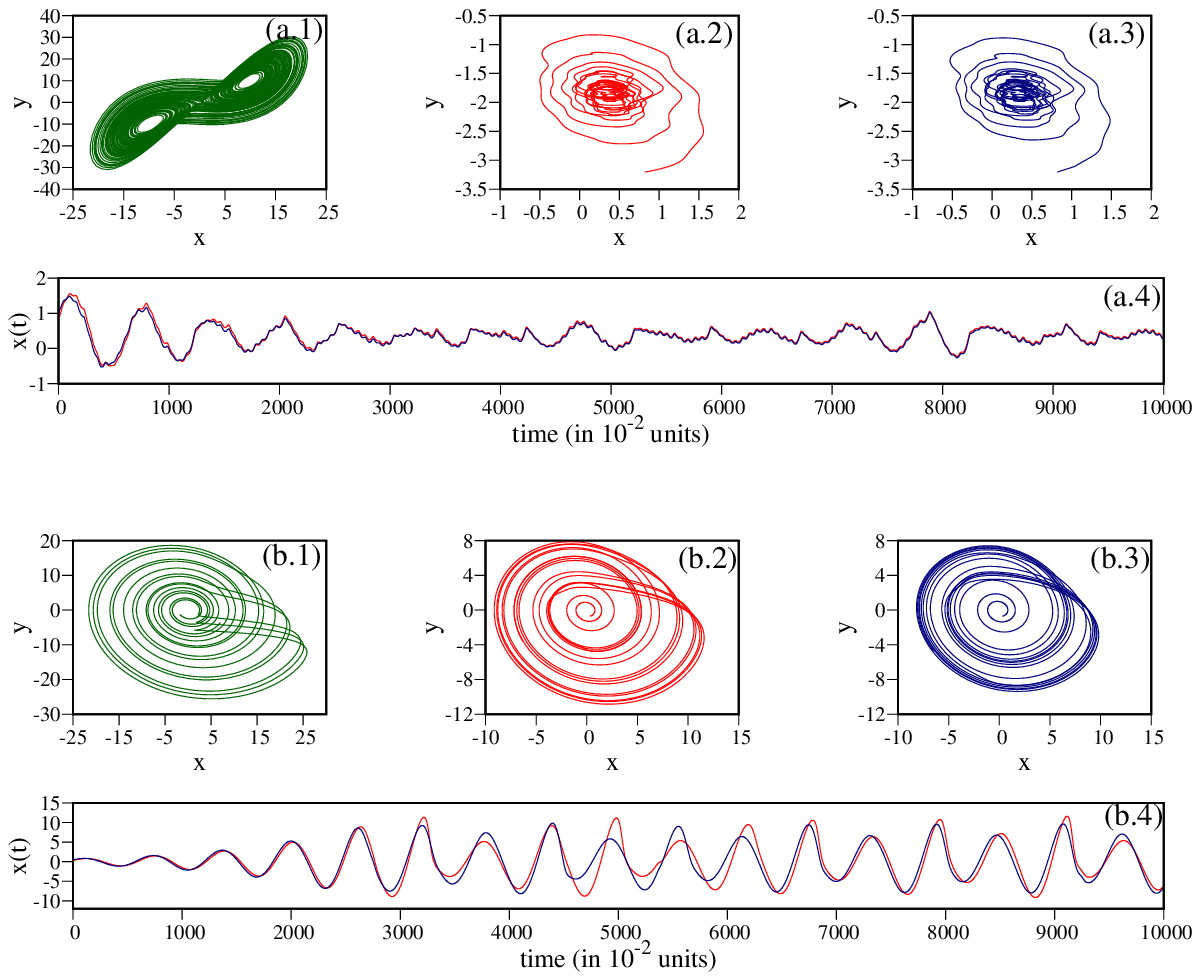}
    \caption{Results of ESN's prediction while trained with Lorenz-R{\"o}ssler system. (a.1) Lorenz drive signal for $\rho_{lor} = 38$, (a.2 - a.3) respectively, the target and machine prediction for response dynamics corresponding to (a.1), (a.4) comparison of target (red line) and predicted (blue line) time series with $\rm{log_{10}(NRMSE)} = -4.58$.
    (b.1) R{\"o}ssler drive signal for $c'_{ros} = 14$, (b.2 - b.3) respectively, the target and machine prediction for response dynamics corresponding to (b.1), (b.4) comparison of target (red line) and predicted (blue line) time series with $\rm{log_{10}(NRMSE)} = -4.03$.}
    \label{fig:res2}
\end{figure*}

In the second case, we train the machine with Lorenz-R{\"o}ssler system given in Eq.~(\ref{eq:lor-ros}) for three different dynamics corresponding to $\rho_{lor} \in [28,32,36]$. Then we check the prediction performance for Lorenz drive signal at $\rho_{lor} = 38$ and R{\"o}ssler drive signal at $c'_{ros} = 14$ as depicte in Fig.~\ref{fig:res2}. For this simulation, we use the Bayesian optimal hyper-parameters as $\sigma = 0.0639$, $\rho = 0.5057$, $\alpha = 0.0409$ and $\beta = 7.026\times 10^{-8}$.

For both cases, coupling strength ($\epsilon$) is kept the same for training and prediction. For all the results presented in this article, $\epsilon = 0.3$. The training length is taken to be $60000$ time steps of the dynamics while the corresponding model is solved with a time step length $0.01$. At the time of training with each time series, the initial $1000$ reservoir evolution states have been discarded as transient. Reservoir states corresponding to all three training time series are used in a single sequence for the regression, stacking one above another. The label (target output) sequence for training is also prepared accordingly.

\begin{figure}
    \centering
    \includegraphics[scale=0.7]{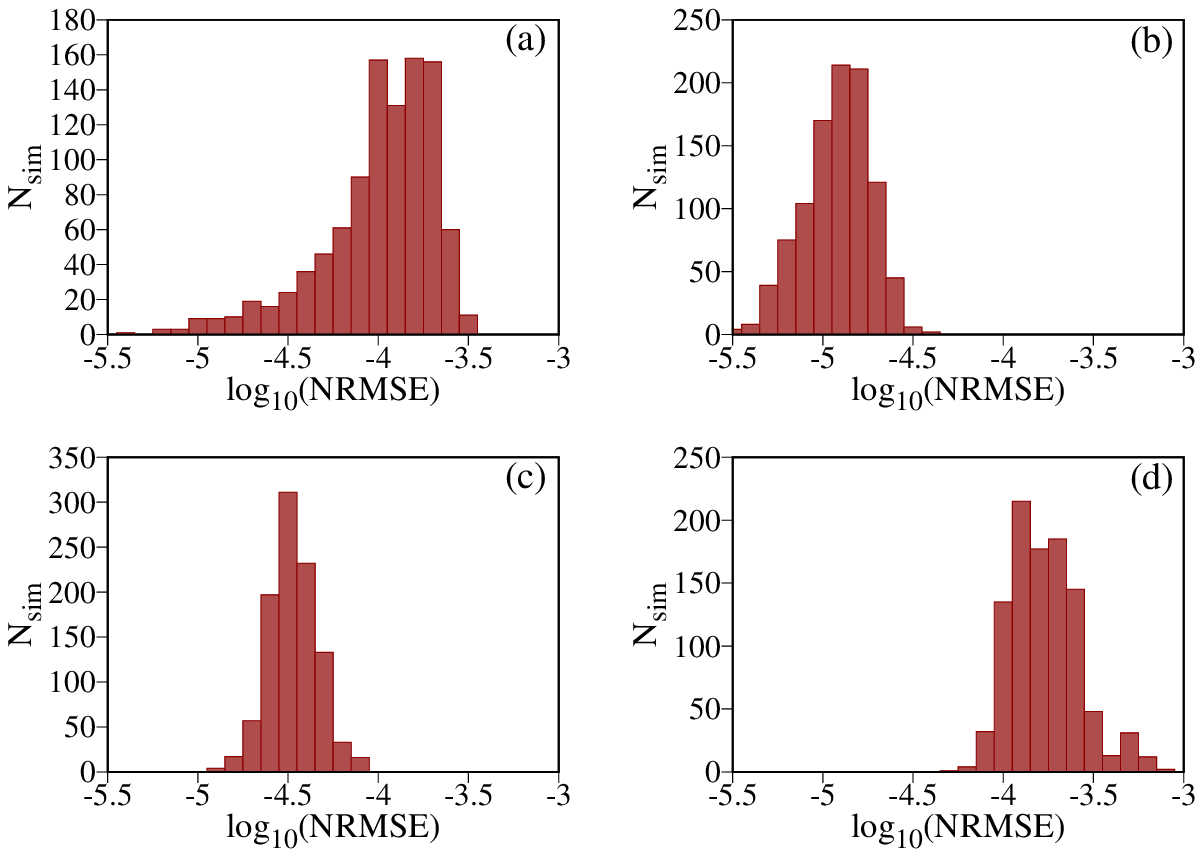}
    \caption{Plot of number of simulations ($\rm{N}_{\rm{sim}}$) Vs accuracy, $\rm{log_{10}(NRMSE)}$ with histograms. (a)-(d) corresponds to predictions for multiple simulations for cases depicted in Fig.~\ref{fig:res1}(a.4), Fig.~\ref{fig:res1}(b.4), Fig.~\ref{fig:res2}(a.4) and Fig.~\ref{fig:res2}(b.4) respectively.}
    \label{fig:rmse}
\end{figure}

Both Fig.~\ref{fig:res1} and Fig.~\ref{fig:res2} show the performance of the ESN for one random reservoir realization. Clearly, the results depict a notable success of our scheme. Though the forecast of time series starts deviating after a certain time, the machine produces the nature of the attractor pretty well. However, a quantitative description of the performance is given in Fig.~\ref{fig:rmse}, in terms of the normalized root-mean-square error (NRMSE) in predicting a time series for 10000 time steps.
To show the robustness in performance, we take 1000 realizations of the reservoir and note the prediction accuracy in each case. The histogram of simulation number ($\rm{N_{sim}}$) with the accuracy ($\rm{log_{10}NRMSE}$) shows a high-quality factor for all cases as the evidence of robust performance of the scheme.

Table~\ref{tab:scale} depicts the generality of the results presented. The tabulated numbers refer to $\rm{log_{10}NRMSE}$ values of response predictions for a Lorenz drive signal ($\rho_{lor} = 38$) and its scaled forms. It was scaled both in amplitude and frequency by upto $\pm30\%$. It shows that, the prediction accuracy of the ESN does not suffer significantly even for a scaled version of the new drive signal. With this, one can argue that the choice of a unknown driving signal to verify the machine's learning can be quite general.

\begin{table}[]
\caption{Comparison of performances of the reservoir in terms of the accuracy ($\rm{log_{10}NRMSE}$) of response prediction when the driving signal has been scaled in amplitude as well as frequency with different scaling factors.}
\resizebox{\columnwidth}{!}{%
\tiny 
\begin{tabular}{lcccc}
\hline\hline
                                   &                & \multicolumn{3}{c}{Amplitude Scaling} \\ \cline{3-5} 
                                   & Scaling Factor & 0.7          & 1.0       & 1.3       \\ \hline 
\multirow{3}{*}{Frequency Scaling} & 0.7            & -4.57        & -5.29     & -4.82    \\ 
                                   & 1.0            & -4.47        & -5.42     & -4.71    \\ 
                                   & 1.3            & -4.33        & -5.52     & -4.62        \\ \hline\hline 
\end{tabular}%
}
\label{tab:scale}
\end{table}

\section{Conclusion}\label{sec:disc}

To summarize, we have successfully demonstrated the ability of ESN to learn a unidirectional coupling scheme. Moreover, only a few time series data are enough as examples for the machine to approximate the drive-to-response relation. Once trained with a single drive--response system, machine is able to predict the dynamics of the response system for any given drive signal. We have shown the results using two different systems' data as training samples. For both cases, the ESN architecture is capable of learning the drive-to-response relation with great accuracy. For both the cases we analyse the machine's performance while working with same drive system configured at new parameter as well as a completely different system as drive signal. Finally, we present the quantitative validation for robustness of the reservoir's performance.

In many applications, ESN have been proven to be exceptionally capable including long-term prediction of chaotic signals. But in this work we explore ESN's ability not only in reconstructing a self-excited attractors, but it can also predict the dynamics of a driven system without model just by learning the coupling scheme with the driver. This essentially means the reservoir computer can predict unknown the response system dynamics from the known drive signal. At first glance, it may look like the `observer problem' \cite{lu2017reservoir}, but this application of RC has many subtle yet very important differences from the `reservoir observer'. One unique feature of this scheme is that, unlike the observer problem, it predicts a non-autonomous system's dynamics which is driven by a completely independent drive signal. Moreover, the reservoir observer works only for subsequent states of the training data. But in our scheme, once the machine is trained with examples, it can predict the dynamics of a driven system starting from any random initial state just by looking at the drive signal. This implies, in the training phase even from a very few example time series machine was able to learn the actual non-driven (autonomous) dynamics of the system and how it is affected in the presence of a drive signal. Which essentially results in learning the unidirectional coupling scheme as well as the original dynamics of the system when not driven. The scheme presented in this article is also quite general and can be applied to varied types of systems and coupling schemes. We have checked its performance for the prediction of a driven Lorenz system too (results are not presented). In that case, also, the machine performs well. But due to the extremely chaotic nature of the predicting system, the length of the predicted trajectory is shorter for a given accuracy. The scheme also has been verified to work with non-linear coupling as well.
Also, it is capable of learning the coupling even when the systems are coupled uni-directionally with more than one state variable.

However, there are some limitations of this scheme too. In cases with multistability in the training data set, the ESN seems to fail in learning the coupling properly. The accuracy of prediction significantly dropped for those cases. Also, the scheme loses its performance while we swap the drive system with an alternate one after training in the case of a multi-dimensional driving signal from a single system. This is due to the reservoir's observer property as it establishes a relation between two drive signals which does not hold for drive signals from a different system. This leads to an ambiguity resulting a dip in the performance. Moreover, the scheme seems to work only for drive signals from deterministic systems, it becomes inapplicable for a random stochastic drive signal. Exploring these issues in detail may be a scope for future studies in this area of reservoir computing applications. Additionally, this study can be extended further to evaluate the exact mathematical forms of the coupling.

In conclusion then, this article demonstrates the remarkable ability of ESN for data-driven prediction of a driven system by learning the unidirectional coupling with its driver. The results of this work opens up many dimensions for research regarding exploration of coupled oscillators' behavior exploiting reservoir computing. There exist numerous natural dynamical with intentional or unintentional presence external signals like environment and noise which influence the original dynamics of the system. In many cases, it is difficult to model such system's dynamics. The scheme presented here can be a great tool for such cases. It can learn the external signal's effect on the original system and help to analyse its dynamical behavior. Thus, this work take a significant step forward in studying complex dynamical systems utilizing modern tools of machine learning.

\begin{acknowledgements}
Authors acknowledge financial support from SERB, Department of Science and Technology (DST), India (Grant No. CRG/2021/003301). 
\end{acknowledgements}

\bibliography{citations}

\end{document}